\pdfoutput=1
\documentclass[11pt]{article}

\usepackage[]{acl}

\usepackage{times}
\usepackage{latexsym}
\usepackage{graphicx}
\usepackage{amssymb}
\usepackage{amsmath}
\usepackage{mathtools, nccmath}
\usepackage{tabularx}         % fancy tables
\usepackage{booktabs}         % fancy tables
\usepackage{multirow,multicol}         % fancy tables
\usepackage{cuted}
\usepackage[T1]{fontenc}
\usepackage[utf8]{inputenc}

\usepackage{microtype}

\title{Auto-Select Reading Passages in English Assessment Tests?}

\author{Bruce W. Lee$^{1,2}$ \\
  University of Pennsylvania$^{1}$ \\
  \texttt{brucelws@seas.upenn.edu} \\\And
  Jason Hyung-Jong Lee$^{2}$ \\
  LXPER AI Research$^{2}$ \\
  \texttt{jasonlee@lxper.com} \\}

\begin{document}
\maketitle
\begin{abstract}
We show a method to auto-select reading passages in English assessment tests and share some key insights that can be helpful in related fields. In specifics, we prove that finding a \textbf{similar} passage (to a passage that already appeared in the test) can give a \textbf{suitable} passage for test development. In the process, we create a simple database-tagger-filter algorithm and perform a human evaluation. However, 1. the textual features, that we analyzed, lack coverage, and 2. we fail to find meaningful correlations between each feature and suitability score. Lastly, we describe the future developments to improve automated reading passage selection.
\end{abstract}

\section{Introduction}
\subsection{Why Important?}
Not all reading materials apply the same. Depending on how the reading material is presented to language learners, generally good reading material might not be as appropriate in more specific contexts. In this paper, we investigate if automatic reading passage selection is a feasible task in test-based contexts.

Though not many, some studies \citep{kuo2014designing, hsu2010development} attempted to devise a reading material recommendation system to give generally helpful texts to students. Such systems attempt to give what is helpful to each user, based on one's performance record. However, most English assessment exams (SAT, TOEFL, TOEIC) are standardized, presenting some unique challenges. 

Choosing appropriate passages for reading comprehension sections of such exams is a lengthy process. SAT, a standardized test widely used for college admissions in the United States \citep{westrick2019validity}, has lengthy technical specifications for test development, which include three steps: 1. item search/crafting, 2. content and fairness review, 3. pretesting \citep{college2017sat}. Each step requires further technical training of already professionals, making test development troublesome. 

The 1. test-specific technicalities merge with the 2. test developer's expertise to select suitable reading passages. Intuitively speaking, both components are difficult to mimic even with the recent natural language processing (NLP) techniques. Hence, automatic reading passage selection for test papers has largely been left in the dark. But the topic is worthy enough of investigation.

Faster reading passage selection is definitely useful as it enables cheaper testing, diverse variations of test papers, better student preparation, etc. But would deeper research produce a usable passage selection system for tests? We try to answer.

\subsection{Find Alike, not Find Again}
We propose a multi-aspect text similarity search on a gold instance, instead of a bottom-up recommendation algorithm. A gold instance is a reading passage that has already appeared in previous test papers. Therefore, a gold instance has already gone through a standard passage selection procedure. 

By searching for a text similar to the gold instance, instead of mimicking the full selection procedure, we can utilize the textual properties of the gold and skip steps difficult to imitate.

We need a multi-aspect text similarity search because the textual properties that make a passage appropriate are manifold \citep{college2017sat, ets}. Though previous text similarity studies mostly focus on semantic similarity \citep{prakoso2021short, gomaa2013survey}, we need to expand the scope to surface-level properties, lexical difficulty, etc. 

\subsection{College Scholastic Ability Test (CSAT)}
CSAT tests the ability to study in college, based on South Korea's high-school curriculum \citep{csat}. Around 500,000 Korean students take the test every year. English is a required subject, and it puts much focus on reading comprehension. Each test paper has more than 20 reading passages taken from high-quality, previously published sources. We take CSAT as a running example.

\section{Model Description}
\subsection{Overview}
Our proposed system consists of three main components: 1. \textit{database}, 2. \textit{extractor}, 3. \textit{filter}. A database is a collection of stand-alone passages. The database is an arbitrary concept; it could be a gold instance's original published source (considering a gold is an excerpt), the ACL Anthology, or even the whole internet. Developing a database is analogous to setting the system's sensible search range. The passages in a database are run through extractors to give numerical representations of textual features.

Extractors are a collection of multiple single feature extractors. Some examples are a \textit{word per sentence extractor} and a \textit{semantic encoding extractor}. The numerical features (integer, float, or multiple numeric sequences), are then concatenated to the original passages to be saved in the database, awaiting for a gold instance to be passed.

When a gold instance\footnote{In this paper, "instance" is used analogously to "passage".} enters the system (Figure 1), it is first run through the same set of extractors (but not saved in the database). Then, the whole database goes through multiple filters to abandon dissimilar passages. The filters have \textit{pass conditions} that compare the corresponding features between the database passages and gold instance's to select a passage appropriate to replace gold.

\begin{figure}
    \includegraphics[width=0.5\textwidth]{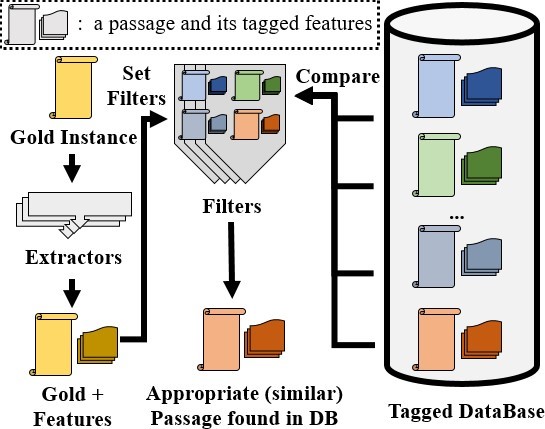}
    \caption{Our system. Each color is a unique passage.}
    \vspace{-4mm}
\end{figure}

\subsection{Database}
\begin{table}
\centering
\resizebox{0.5\textwidth}{!}{%
\begin{tabular}{lccccc}
\cmidrule(lr){1-6}
\textbf{Properties}    & \textbf{\#Instance} & \textbf{\#Sent}  & \textbf{\#Word} & \textbf{\#Char} & \textbf{\#Diff}\\ 
\cmidrule(lr){1-1}\cmidrule(lr){2-2}\cmidrule(lr){3-3}\cmidrule(lr){4-4}\cmidrule(lr){5-5}\cmidrule(lr){6-6}
total                        & 1,044     & 10,651  & 2.2e5        & 1.4e6    & 23,792    \\
mean                          & N/A       & 10.2    & 215          & 1,339    & 22.8    \\
Q4 (max)                      & N/A       & 37      & 431          & 2,802    & 70    \\
Q3                            & N/A       & 12      & 254          & 1,588    & 28    \\
Q2 (median)                   & N/A       & 10      & 212          & 1,322    & 22       \\
Q1                            & N/A       & 7       & 173          & 1,078    & 16    \\
min                           & N/A       & 1       & 7            & 42       & 0    \\
\cmidrule(lr){1-6}
\end{tabular}
}
\caption{\label{Table 1} Database statistics. \#: number of. diff: words that only appear grade 9 and above.}
\vspace{-4mm}
\end{table}

Our database is a collection of instances, mostly a single paragraph, like in CSAT. All past CSAT English subject papers are publicly available\footnote{www.kice.re.kr}. We take the test papers between 2021-2022; we searched the source articles.

We used optical character recognition (OCR) through Google Vision API on the retrieved source articles' image files. We manually divided the result into paragraphs but we did not perform any inspection on the appropriateness of each item. We only went through basic preprocessing procedures. The database distribution is described in Table 1. The preprocessing steps are:

\noindent \textbf{1.} remove repeating whitespace characters

\noindent \textbf{2.} remove sentences of lengths under 10 characters

\noindent \textbf{3.} remove passages of lengths under 30 characters.

\subsection{Extractor}
Then, we develop a series of extractors. The first is sentence-word counter using \textit{en\_core\_web\_sm} model pipeline from spaCy v.3.2.1 \citep{spacy2}. We pass in each passage into a spaCy object and count items in the resulting token list and sentence list. Usages are publicly available\footnote{spacy.io/usage/spacy-101}.

Second, we implement a lexical difficulty extractor using the Word Corpus of the Korean ELT curriculum (CoKEC-word) from \citet{lee-lee-2020-lxper}. The dataset classifies 30608 words in 6 levels, with each corresponding to a specific grade range  (A: suitable for grade 1 to 4 students, B: grade 5 to 8 students, C: grade 8 to 9 students, D: grade 9 to 11 students, E: grade 11 to 12 students, F: college).

Last, we implement semantic embeddings extractor using SBERT-based framework \citep{reimers-gurevych-2019-sentence}. From the authors' \textit{sentence-transformers} library, we utilize the recommended model (\textit{all-MiniLM-L6-v2}) \citep{wang2020minilm}.     

\subsection{Filter}
A filter is a function of some \textit{data} and a \textit{condition} (\textit{filtered data = filter(data, condition)}). Whereas a condition is a function of a gold instance, a database instance, and a parameter $\alpha$ (\textit{pass/fail = condition(gold instance, DB instance, parameter)}). 

\begin{table}
\centering
\resizebox{0.5\textwidth}{!}{%
\begin{tabular}{lcccccc}
\cmidrule(lr){1-7}
\textbf{Gold}    & \textbf{\#Sent} & \textbf{WPS}  & \textbf{H\_diff} & \textbf{HPW} & \textbf{Sem. Similar} & \textbf{\#Output}\\ 
\cmidrule(lr){1-1}\cmidrule(lr){2-2}\cmidrule(lr){3-3}\cmidrule(lr){4-4}\cmidrule(lr){5-5}\cmidrule(lr){6-6}\cmidrule(lr){7-7}
1                     & 2         & 10      & 0            & 0.1      & 0.35    & 9    \\
2                     & 2         & 10      & 0            & 0.1      & 0.35    & 7    \\
3                     & 2         & 10      & 0            & 0.1      & 0.37    & 5    \\
4                     & 4         & 10      & 1            & 0.05     & 0.35    & 5    \\
5                     & 4         & 10      & 0            & 0.05     & 0.35    & 5     \\
6                     & 4         & 10      & 0            & 0.05     & 0.33    & 5     \\
7                     & 2         & 10      & 0            & 0.05     & 0.22    & 6     \\
8                     & 2         & 10      & 0            & 0.05     & 0.22    & 6     \\
9                     & 2         & 10      & 0            & 0.05     & 0.35    & 5     \\
10                    & 2         & 10      & 0            & 0.05     & 0.33    & 6     \\
\cmidrule(lr){1-7}
\end{tabular}
}
\caption{\label{Table 2} Parameters $\alpha$ set for each gold instance.}
\vspace{-4mm}
\end{table}

Combining multiple filters in a sequence reduces the original database multiple times to leave only the passages that pass all conditions under given parameters. Below are the five conditions we use, in a consecutive manner from top to bottom. 

\vspace{4mm}

\noindent \textbf{Sentence count condition}:
\vspace{-4mm}
\begin{equation*}
\text{G's \#Sent} - \alpha \leq \text{D's \#Sent} \leq \text{G's \#Sent} + \alpha
\end{equation*} 

\noindent \textbf{Word per sentence condition}:
\vspace{-4mm}
\begin{equation*}
\text{G's WPS} - \alpha \leq \text{D's WPS} \leq \text{G's WPS} + \alpha
\end{equation*} 

\noindent \textbf{Max word difficulty condition}:
\vspace{-4mm}
\begin{equation*}
\text{G's H\_diff} - \alpha \leq \text{D's H\_diff} \leq \text{G's H\_diff} + \alpha
\end{equation*}

\noindent \textbf{Difficult word per all word condition}:
\vspace{-4mm}
\begin{equation*}
\text{G's HPW} - \alpha \leq \text{D's HPW} \leq \text{G's HPW} + \alpha
\end{equation*}

\noindent \textbf{Semantic similarity condition}:
\vspace{-4mm}
\begin{equation*}
\alpha \leq \text{G and D's Semantic Similarity} 
\end{equation*}

For the above conditions, \#Sent: number of sentences, WPS: \#word per sentence, H\_diff: highest word difficulty, HPW: \#high difficulty word (grade 9 and above) per word. $\alpha$ is the parameter that sets the range. Semantic similarity is calculated through the cosine distance between gold instance's (G) semantic encoding and database instance's (D).

\section{Evaluation}
\subsection{Test Arrangement}
We randomly selected 10 gold instances from CSAT papers between 2021 and 2022. Since all CSAT reading passages are of similar lengths, the selected passages ranged from 134 to 181 words. From our database (Section 2.2), we ran the conditions (Section 2.4) to obtain 5 candidate reading passages for each gold instance.

The parameters we set for each condition are described in Table 2. These conditions are manually fine-adjusted to make the resulting number of instances as close to 5. As shown in the \#Output column, it sometimes seemed impossible to give the desired \#Output. This is due to the statistical distribution of the respective features in database.

The 5 instances that passed our conditions were concatenated to 5 dummy instances, from the same database but completely randomly selected. The 10 instances were shuffled for anonymity. Ultimately, we obtained 100 candidate instances along with 10 gold instances.

We presented the test set to 2 EFL teachers in South Korea, each with over 10 years of experience, and 1 South Korean EFL student who took CSAT in 2020. All 3 test subjects are employees in EFL educational firms or institutions.

For each trial, 10 candidate instances were shown along with the corresponding gold instance. The test subjects were asked to give suitability scores on a 5-point scale. Our question was "Is the candidate instance suitable as a CSAT reading passage?" They were also asked to give reasons as to why they gave the score.

\begin{table}
\centering
\resizebox{0.5\textwidth}{!}{%
\begin{tabular}{lcccccc}
\cmidrule(lr){1-7}
\multirow{2.4}{*}{\textbf{Gold}} & \multicolumn{3}{c}{\textbf{$\sum$ Score}} & \multicolumn{3}{c}{\textbf{Pearon Correlation}}\\
\cmidrule(lr){2-4}\cmidrule(lr){5-7}
& \textbf{Dummy} & \textbf{Filtered} & \textbf{$\Delta$} & \textbf{\#Word} & \textbf{\#Sent} & \textbf{CEFR}\\ 
\cmidrule(lr){1-1}\cmidrule(lr){2-2}\cmidrule(lr){3-3}\cmidrule(lr){4-4}\cmidrule(lr){5-5}\cmidrule(lr){6-6}\cmidrule(lr){7-7}
1                     & 13         & \textbf{17}    & +4  & 0.102             & -0.0396  & $\sim$0   \\
2                     & 11         & \textbf{16}    & +5  & -0.111            & 0.360    & $\sim$0   \\
3                     & 12         & \textbf{15}    & +3  & -0.336            & -0.859   & $\sim$0   \\
4                     & 9          & \textbf{15}    & +6  & 0.0406            & -0.339   & $\sim$0   \\
5                     & 9          & \textbf{13}    & +4  & 0.0922            & 0.545    & $\sim$0   \\
6                     & 15         & \textbf{18}    & +3  & -0.415            & -0.439   & $\sim$0   \\
7                     & 12         & \textbf{17}    & +5  & -0.105            & 0.338    & $\sim$0   \\
8                     & \textbf{16}& 13             & -3  & -0.057            & 0.084    & $\sim$0   \\
9                     & \textbf{20}& 14             & -6  & 0.0707            & 0.0991   & $\sim$0   \\
10                    & 11         & \textbf{17}    & +6  & -0.143            & -0.0869  & $\sim$0   \\
Average               & 12.8       & \textbf{15.5}  & +2.7& -0.0862           & -0.0337  & $\sim$0   \\
\cmidrule(lr){1-7}
\end{tabular}
}
\caption{\label{Table 3} Test results, in terms of score and correlation.}
\vspace{-4mm}
\end{table}

\subsection{Analysis \& Key Insights}
We are interested in seeing if the filtered instances score meaningfully higher than the dummy instances. Table 3 shows the result score sums and correlations of the test outlined in Section 3.1. The Common European Framework of Reference (CEFR) levels is classified with the RoBERTa-RF-T1 model, retrieved from \citet{lee-etal-2021-pushing}. The model is state-of-the-art and gives 76.3\% accuracy on the Cambridge dataset \citep{xia-etal-2016-text}.

\textbf{1. Multi-aspect similarity search gives meaningful statistical improvement from random selection.} The filtered passages triumphed dummy passages in terms of suitability as a CSAT reading passage. This is an interesting phenomenon as the filters were not set to directly measure suitability. Rather, they measure similarity in terms of surface-level, lexical, and semantic textual features. 

This hints that finding a passage \textbf{similar} to a gold instance has the potential to give a passage \textbf{suitable} for test papers. The filters used were relatively simple; with further improvements, the performance can improve. However, we failed to find meaningful correlations with any of the specific features. This leads us to our second insight.

\textbf{2. "Suitability" is a multi-aspect textual property, definitely not captured on the surface level.} From Table 3 and Fig. 2, it seems clear that one does not think a passage is suitable just by \#words or \#sentences. In addition, the classified CEFR level of a passage had no significant correlation with its suitability. The CEFR levels are analogous to a passage's readability \citep{xia-etal-2016-text}, which seems to have minimal correlation with suitability. 

In fact, we find that no meaningful correlation and distribution exist between any tested single feature and suitability. On how filtered passages scored better than dummy passages, we believe that certain combinations of textual features lead to high suitability as a test reading passage.

One expects to see a certain \textit{type} of a passage in a test paper. Such suitability is not measured by objective aggregation of textual features, unlike some other NLP tasks like readability assessment \citep{collins2014computational}. Hence, common prediction models like regression would not work well, complicating the task solution.

\begin{figure}
    \includegraphics[width=0.5\textwidth]{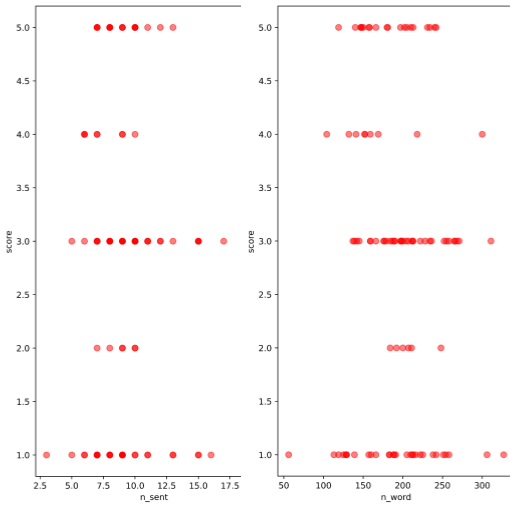}
    \caption{Scatter plots showing unclear trends. y label: \textbf{suitability score}, x label: \textbf{\#sent} (left) and \textbf{\#word} (right).}
    \vspace{-4mm}
\end{figure}    

\section{Weaknesses \& Future Directions}
\subsection{Contextual Dependency}
For a passage to be appropriate for reading comprehension tests, it is required that the passage is understandable by itself. The passage could be self-contained and not dependent on the preceding text, table, or figure. Otherwise, the uncertain parts should be easily inferable. Along with more sophisticated preprocessing methods that detect strings like "Table" or "Figure", we plan to develop inter-paragraph dependency detection algorithms.

\subsection{Evaluation Method}
The test outlined in Section 3 is based on a small population. Further, it is human-dependent and is not easily reproducible. In future works, we plan to introduce an objective evaluation method. This possibly consists of a classification dataset, where gold instances are mixed with its neighboring paragraphs from the same source article. Then, a model could be adjusted to choose the gold instance among the neighboring paragraphs.

\subsection{Auto-Calibration of Filter Conditions}
In Section 3.1, we manually set the parameters $\alpha$ for each condition. This is yet another weakness that makes the study difficult to reproduce. For our filter-based method to be widely useful for both commercial and academic purposes, we need an algorithm that automatically adjusts condition parameters, based on the number of desired outputs.

\section{Conclusion}
We proved that finding a similar passage (to an already test-appeared passage, or a gold instance) has the possibility to give a suitable passage for test development. Due to lacking correlation or distributional relationship between any single feature and suitability rating, we could not prove why we obtained our desired outcome. But we postulated that certain combinations of textual features lead to high suitability as one expects to see certain types of passages in a test paper.

\bibliography{anthology,custom}
\bibliographystyle{acl_natbib}

\appendix
\section{Sample Gold Instance}
--- Mending and restoring objects often require even more creativity than original production. The preindustrial blacksmith made things to order for people in his immediate community; customizing the product, modifying or transforming it according to the user, was routine. Customers would bring things back if something went wrong; repair was thus an extension of fabrication. With industrialization and eventually with mass production, making things became the province of machine tenders with limited knowledge. But repair continued to requirea larger grasp of design and materials, an understanding of the whole and a comprehension of the designer’s intentions. “Manufacturers all work by machinery or by vast subdivisionof labour and not, so to speak, by hand,” an 1896 Manual of Mending and Repairing explained. “But all repairing must be done by hand. We can make every detail of a watch or of agun by machinery, but the machine cannot mend it when broken, much less a clock or a pistol!” ---

The above (gold instance) is from 2022 CSAT, <Waste and Want: A Social History of Trash>. Henry Holt and Company. Susan Strasser. 2014

\end{document}